\documentclass[fullpaper]{nldl}

\renewcommand{\DOI}{12345}

\usepackage[utf8]{inputenc}
\usepackage{url}
\usepackage{graphicx}
\usepackage{authblk}

\usepackage{algorithm}
\usepackage{algorithmic}

\usepackage{caption}
\usepackage{subcaption}
\usepackage{float}
\usepackage{amssymb}
\usepackage{amsmath}
\usepackage{doi}
\usepackage{url}

\usepackage{hyperref}

\title{Uncertainty Quantification of Surrogate Explanations: an Ordinal Consensus Approach}
\author[1]{Jonas Schulz}
\author[2]{Rafael Poyiadzi}
\author[2]{Raul Santos-Rodriguez}

\affil[1]{TU Dresden}
\affil[2]{University of Bristol}

\date{\vspace{-5ex}}

\begin{document}
\nldlmaketitle

\begin{abstract}  
  Explainability of black-box machine learning models is crucial, in particular when deployed in critical applications such as medicine or autonomous cars. Existing approaches produce explanations for the predictions of models, however, how to assess the quality and reliability of such explanations remains an open question. In this paper we take a step further in order to provide the practitioner with tools to judge the trustworthiness of an explanation. To this end, we produce estimates of the uncertainty of a given explanation by measuring the ordinal consensus amongst a set of diverse bootstrapped surrogate explainers. While we encourage diversity by using ensemble techniques, we propose and analyse metrics to aggregate the information contained within the set of explainers through a rating scheme. We empirically illustrate the properties of this approach through experiments on state-of-the-art Convolutional Neural Network ensembles. Furthermore, through tailored visualisations, we show specific examples of situations where uncertainty estimates offer concrete actionable insights to the user beyond those arising from standard surrogate explainers.
\end{abstract}

\section{Introduction}
Deep learning models are being used in critical applications which demand not only human oversight but for their predictions to be explained as well, if they are to be considered \textit{trustworthy}. Explainability tools have been developed aiming to make black-box classifiers interpretable~(\cite{arya2019explanation,kacper2020}). Surrogate explainers, such as Local Interpretable Model-agnostic Explanations (LIME) \cite{lime}, provide an explanation by fitting an interpretable surrogate model to explain the prediction of an instance.

However, explanations produced by LIME can vary due to the hyperparameters of the procedure. Several papers have looked into the shortcomings of LIME and proposed more robust versions~(\cite{blimey,hepburn2021}). In particular, the main sources of uncertainty affecting LIME explanations are studied in \cite{trustlime}. Their work analyses the uncertainty due to LIME's hyperparameters, and also the stochasticity in the process of generating the explanation. 
Similarly, \cite{gradlime} presents both a theoretical and an empirical analysis of the variability of explanations produced for a single image. These results suggest that the inherent stochasticity of LIME induces diversity among multiple explanations produced for the same instance. The idea of applying LIME multiple times to an instance is proposed in \cite{polishpaper}, while the robustness of LIME, with regards to changes in the input data, is explored in \cite{limerobustness}. 


In this paper, we take a step back and aim to enrich the explanations by incorporating an estimate of their uncertainty. This allows for a more meaningful interaction, potentially enabling the user to either trust or reject the explanation
Our contributions are as follows:
\begin{enumerate}
\item We provide uncertainty estimates for explanations using bootstrapping and ordinal consensus metrics. We showcase these using tailored visualisations that convey this information for the practitioner.
\item Beyond the uncertainty within LIME and the uncertainty induced by the input data, we also consider the predictive uncertainty. We do this by considering the model of interest to be an ensemble of black-box models, rather than a single black-box. 
\item We highlight the number of surrogates and the number of instances the surrogates are fitted to as key parameters that help the user fine-tune and adjust our proposed procedure depending on the use case
\end{enumerate}

\section{Related Work}\label{relatedwork}
The process of deriving surrogate explainers is complex and driven by several interconnected factors and objectives (\cite{poyiadzi2021understanding, poyiadzi2021overlooked}). In general, this type of explainers can be unstable and lead to varying surrogate coefficients and, in consequence, diverse explanations (\cite{limerobustness, trustlime, s-lime}). The variability within surrogate coefficients can be seen as uncertainty that surrogate explanations are entailed with. While \cite{polishpaper} and \cite{gradlime} highlight the sampling space where the surrogate is fitted as a source of uncertainty, \cite{clue} motivates the need of also considering the predictive uncertainty of the black-box to be explained. 

In this work we address the quantification of the surrogate explanation uncertainty by aggregating multiple surrogate coefficients. The use of a consensus mechanism to obtain explanations that are less sensitive to sampling variance (further discussed in Section \ref{background}) has been proposed in \cite{bhatt2019towards, rieger2019aggregating}. Specifically, \cite{bhatt2020evaluating} and \cite{bhatt2019building} consider aggregating surrogate coefficients in the form of simple ranking schemes inspired from the social sciences and economics.

\texttt{BayesLIME} was proposed in \cite{slack2020reliable} to generate surrogate explanations with a measure of uncertainty. The uncertainty is quantified by evaluating the probability that surrogate coefficients lie within their 95\% credible intervals. The work suggests sampling perturbations that yield most information to the models behaviour, thus reducing the computational complexity. The practitioner is informed about the uncertainty of feature attribution to each explainable component. 

The idea of using uncertainty-aware black-box models for interpretability has been investigated in \cite{tromsopaper1} and \cite{tromsopaper2} and demonstrated on various saliency mapping methods. However, little work has been done on combining surrogate explanations with model uncertainty \cite{clue}. In this paper, we introduce a framework that uses the aggregation of multiple diverse surrogate explainers in combination with uncertainty aware deep learning ensembles. Similarly to \cite{slack2020reliable}, we motivate the number of perturbations sampled as well as the number of surrogates derived as key parameters the practitioner needs to fine tune in order to derive explanations that satisfy the desired certainty of the surrogate derivation process.

\section{Background}\label{background}
In this section, we briefly introduce local-surrogate explanations, with a particular focus on LIME (\cite{lime}). We highlight different sources of surrogate uncertainty and discuss two in particular which, as further described in Section \ref{Uncertainty Quantification of Surrogate Explanations}, are used to naturally induce diversity among multiple bootstrapped surrogates.

\subsection{Surrogate Explanations}
Local-surrogate explanations belong to a category of post-hoc model-agnostic explanation approaches first introduced in \cite{lime}. One such approach is LIME, which is an instantiation of the following formulation:
\begin{equation}\label{limeeqn}
    \begin{aligned}
        & \underset{g\in \mathcal{G}}{\arg\min}
        & \mathcal{L}(f,g,\pi_{x})+\Omega(g). 
    \end{aligned}
\end{equation}
The surrogate explainer $g$ is from an interpretable model class $\mathcal{G}$. The locality around the data point $\mathbf{x}$ for which the prediction is to be explained, is controlled by the similarity kernel $\pi_x$. The loss $\mathcal{L}$ characterises how close $g$ is to $f$. The penalty term $\Omega$ represents a complexity measure of $g$. In practice $G$ is the class of linear models: $g(\mathbf{x}) = \boldsymbol{\alpha}^{\top}\mathbf{x}$. Model fitting is performed on a set of points $\mathcal{P}$ drawn from a Gaussian distribution centred on $\mathbf{x}$, and then the weights $\pi_x$ are computed using the radial basis function kernel. 

\begin{figure*}[!t]
\centering
\begin{subfigure}{.49\textwidth}
  \centering
  \includegraphics[width=1\linewidth]{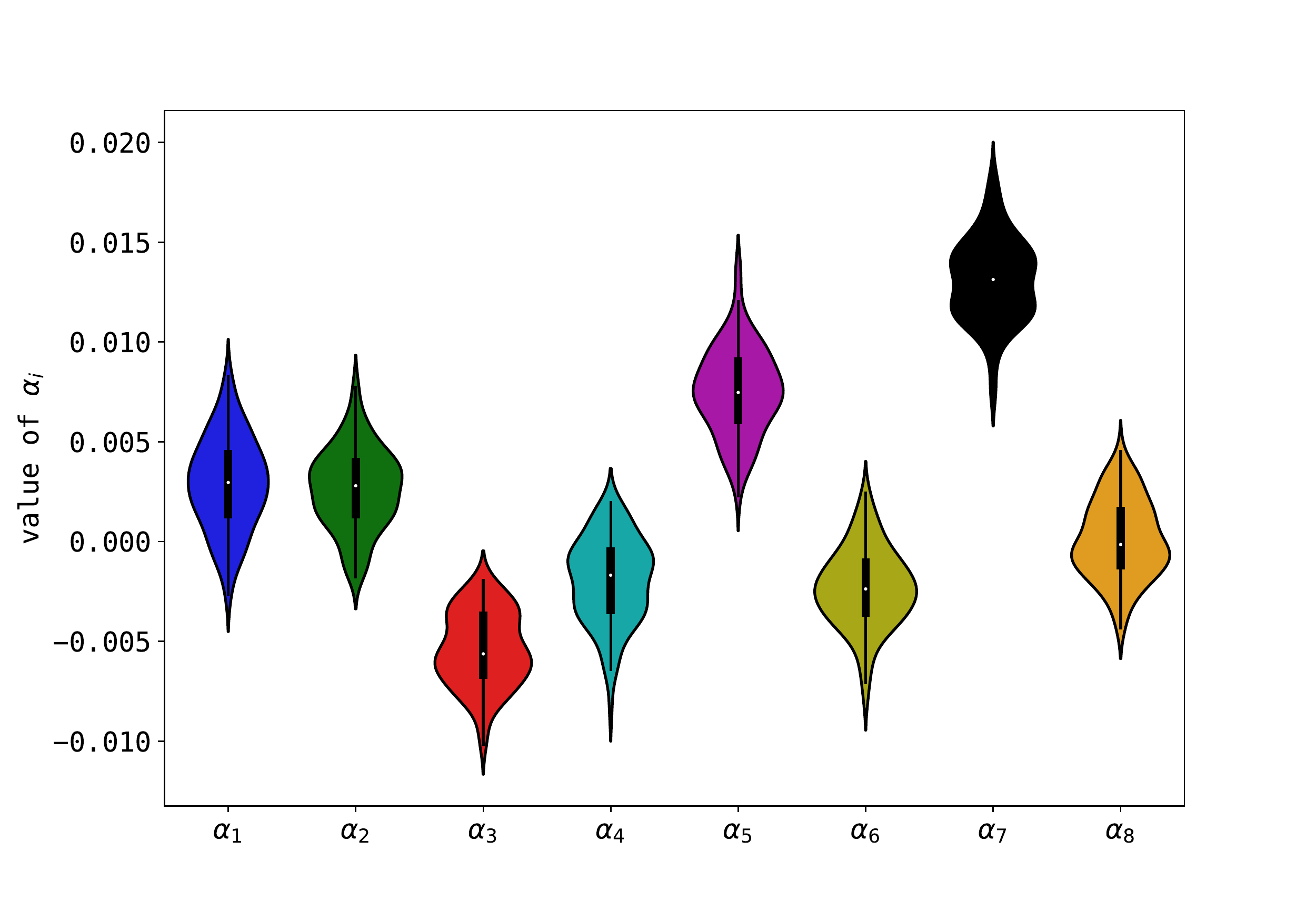}
  \caption{Variability due to sampling variance}
  \label{violin}
\end{subfigure}%
\begin{subfigure}{.49\textwidth}
  \centering
  \includegraphics[width=1\linewidth]{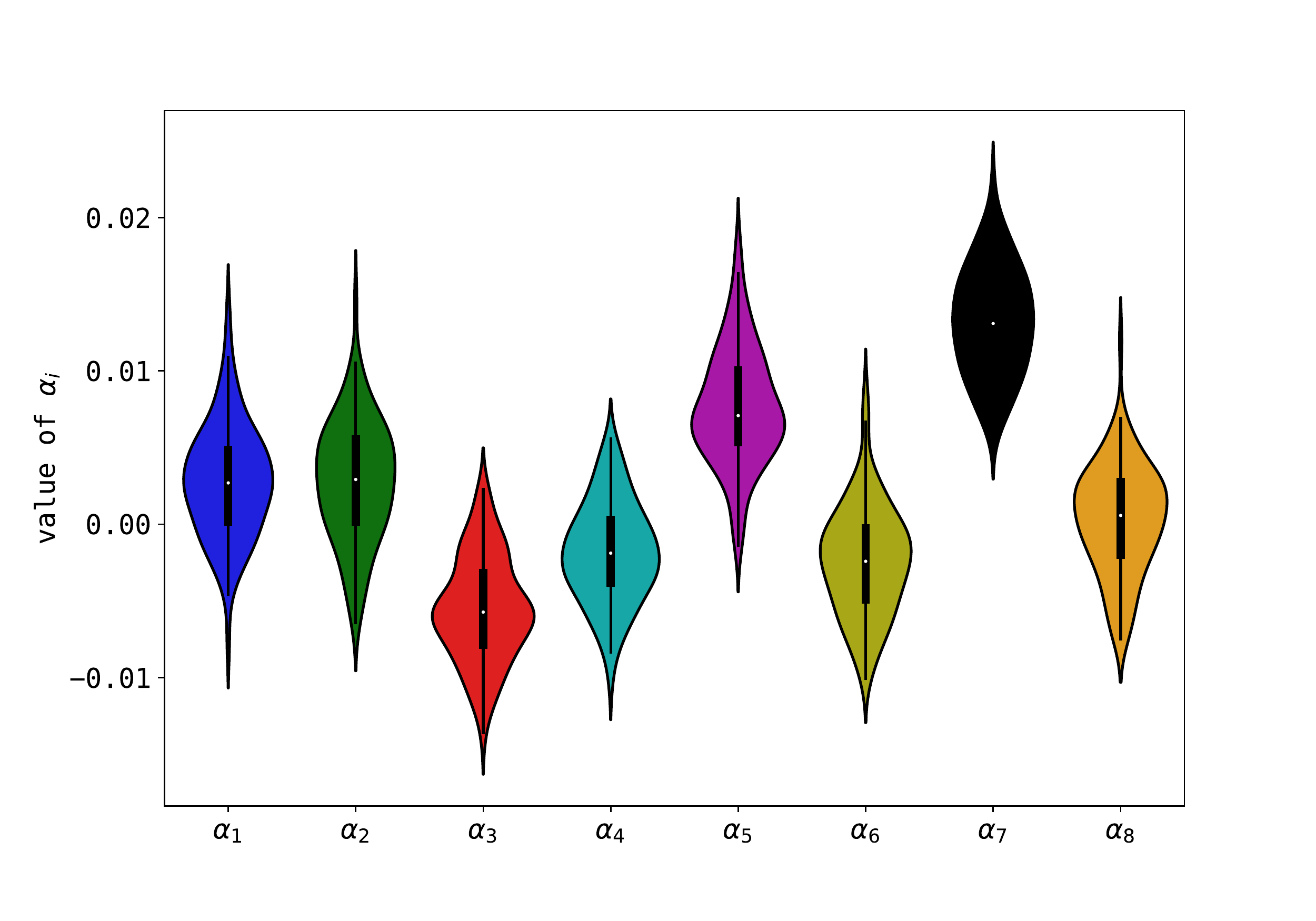}
  \caption{Variability due to predictive uncertainty}
  \label{violin2}
\end{subfigure}
\caption{(a) Distribution of surrogate coefficients $\alpha_i$ derived by LIME on the image depicted in Fig.~\ref{segmentationexample} (left column) on 100 different perturbation sets $\mathcal{P}_k$. (b) Distribution of surrogate coefficients $\alpha_i$ derived by LIME on the image depicted in Fig. \ref{segmentationexample} (left column). LIME is run 100 times with a fixed perturbation set $\mathcal{P}$. The classifier's prediction is sampled randomly from the ensemble members.}
\label{}
\end{figure*}

 
\subsection{Uncertainty in Surrogate Explanations}
Previous works have identified the following sources of uncertainty in surrogate explanations:
\begin{enumerate}
\item \textbf{Sampling variance of $\mathcal{P}$} (examined in \cite{lee2019developing, gradlime, polishpaper, limerobustness}).
\item \textbf{Implementation of explanation procedure} (highlighted in \cite{trustlime}).
\item \textbf{Choice of surrogate structure} (introduced in \cite{trustlime}).
\end{enumerate}
In this paper, assuming sources 2 and 3 fixed, we focus on the sampling variance as well as the predictive uncertainty of the model to be explained. We argue that the predictive uncertainty of the model $f$ can be seen as an extra source of variability, adding to the uncertainty of the explanations. To show this, in the examples below we use ensemble models for their nice properties on uncertainty estimation. Details on the architecture of the ensemble are given in Sec. 5.

\paragraph{Variability of surrogate coefficients due to sampling variance}
\label{variability}
Following the works of \cite{gradlime} and \cite{polishpaper}, Fig.~\ref{violin} shows the variability of surrogate coefficients $\boldsymbol{\alpha}$ due to sampling variance in an image classification task. In images, the explanations are usually based on superpixels given by a segmentation of the image with semantic meaning. Here, LIME is run 100 times (by first drawing 100 distinct sets of points, $\{\mathcal{P}_k\}_{k=1}^{100}$) resulting in 100 surrogates with the default configuration. We generate the predictions by averaging the predictions of the individual ensemble members. Since the sets $\mathcal{P}_k$ of image perturbations are generated randomly, values of $\boldsymbol{\alpha}$ are not deterministic. 

We see that the mean value of $\alpha_7$ is the highest, suggesting that the superpixel $s_7$ can be more clearly identified as, on average, the most relevant region of the image for the classification purposes. $\alpha_3$ can be identified as the least important. For the rest, the ordering is not clear. This is important if the user is interested in tuning LIME such that the distributions of $\alpha_i$ do not overlap so that the order of importance of the coefficients can be clearly identified. 

\paragraph{Variability of explanations due to predictive uncertainty}\label{variabilitypred}
The uncertainty of LIME due to the predictive uncertainty of the black-box classifier has not been addressed in previous works. However, this is something that we can study when using ensemble models. In Fig.~\ref{violin2} LIME is run 100 times on a \textit{fixed set of image perturbations} $\mathcal{P}$ (Compared to above, in this experiment we only have one set of points $\mathcal{P}$, as opposed to above where we use 100 $\{\mathcal{P}_k\}_{k=1}^{100}$). Here, a single prediction is obtained from a randomly chosen member of the ensemble. Therefore, differently from the experiment presented in Fig.~\ref{violin}, the variability of the surrogate coefficients is now solely induced by sampling the predictions $f(\mathbf{x'}_i)$ for image perturbations $\mathbf{x'}_i$ randomly from the individual models contained in the ensemble. Again, for this particular image, the top and bottom coefficient remain the same as before, corroborating the message from the previous example. In Sec.~\ref{results} we will further explore this relationship empirically. In Sec.~\ref{Uncertainty Quantification of Surrogate Explanations}, we present a method of deriving multiple diverse surrogates and aggregate their coefficient values through a rating scheme, allowing to estimate the uncertainty of the aggregated explanation through measures of consensus.

\section{Uncertainty Quantification via Ordinal Consensus}\label{Uncertainty Quantification of Surrogate Explanations}
The coefficients of the surrogate are representative of the behaviour of $f$ locally. A method for estimating the distribution of surrogate coefficients $\boldsymbol{\alpha}$ is bootstrap (\cite{jackknifebootstrap}), as the sampling variance of data points around $\mathbf{x}$ naturally induces diversity among bootstrapped surrogates. Additionally, we propose the use of ensemble techniques to account for the stochasticity of the prediction behaviour of $f$, reinforcing the diversity of the bootstrapped surrogates. We propose ordinal metrics to aggregate the surrogate coefficients and quantify uncertainty. 



\subsection{Bootstrapping LIME}
In our approach, that we refer to as Bootstrapping LIME (BLIME), multiple surrogate models are fitted by bootstrapping the perturbation dataset $\{\mathcal{P}_k\}_{k=1}^K$. Since an ensemble model can be treated as a probabilistic classifier, the output $f(\mathbf{x}'_i)$ for a perturbation $\mathbf{x}'_i$ can also be sampled from the ensemble classifier by sampling a base model from the set of models. 
The BLIME algorithm is as follows. From every surrogate model, we obtain a coefficient vector $\boldsymbol{\alpha}$. Then, for every coefficient vector, we obtain a ranking $\boldsymbol{r}$, ordering coefficients from smallest to largest in value. In this manner, and continuing with the image classification example, if the procedure is repeated $K$ times, for a total of $M$ superpixels, we can compactly represent these ranking vectors as rows of a ranking matrix $\mathbf{R}\in\mathbb{R}^{K \times M}$. $\mathbf{R}$ is then interpreted as a \textbf{rating scheme, where $M$ superpixels are being rated by $K$ surrogates.}

\subsection{Ordinal Metrics}
The reduction of surrogate coefficients to a ranking can be regarded as a normalisation step that makes multiple surrogates comparable. Through ordinal statistics, we can quantify the level of consensus amongst the surrogates to gain further insights for a given explanation. We report the following metrics as proxys to the underlying uncertainty. 

\paragraph{Mean rank} By comparing the mean rank $\overline{r}_j$ of a superpixel $s_j$ to those of all the other superpixels, indicating the relative importance of each of them.
\paragraph{Ordinal consensus} The ordinal consensus $C_j$ of the ranking of a superpixel $s_j$, as defined in \cite{ordinaldispersion}, can be used to evaluate whether there is high agreement among the raters, indicated by $C_j$ being close to 1. A value of $C_j$ closer to 0.5 suggests a rather uniform distribution of rankings assigned to superpixels $s_j$, which means that the importance assigned to $s_j$ varies widely among the surrogates. For $C_j$ closer to zero, ordinal dispersion indicates high polarisation among raters when assigning a rank to $s_j$. 
\paragraph{Inter-rater reliability measures} Here we consider reliability measures to address the overall uncertainty among all surrogates regarding all interpretable components, namely \texttt{Fleiss' Kappa} $\kappa$ \cite{fleisskappa} and \texttt{Kendall's coefficient of concordance} $W$ \cite{kendallw}. While $W$ measures the agreement among raters specifically for rankings, $\kappa$ estimates the agreement  regardless of the similarity of the assigned ranks. 



\section{Experiments}\label{expchapter}
As described in Section \ref{Uncertainty Quantification of Surrogate Explanations}, we propose using the sampling variance of $\mathcal{P}$ and ensemble classifier to induce diversity into bootstrapped surrogates. We identify the number of surrogates aggregated and the size of sets $\mathcal{P}$ as key parameters the user can fine-tune for the explanation derivation process. For the experiments we consider image classification and prediction from text data. 

\begin{figure}
\centering
\begin{subfigure}{.5\textwidth}
  \centering
  \includegraphics[width=1\linewidth]{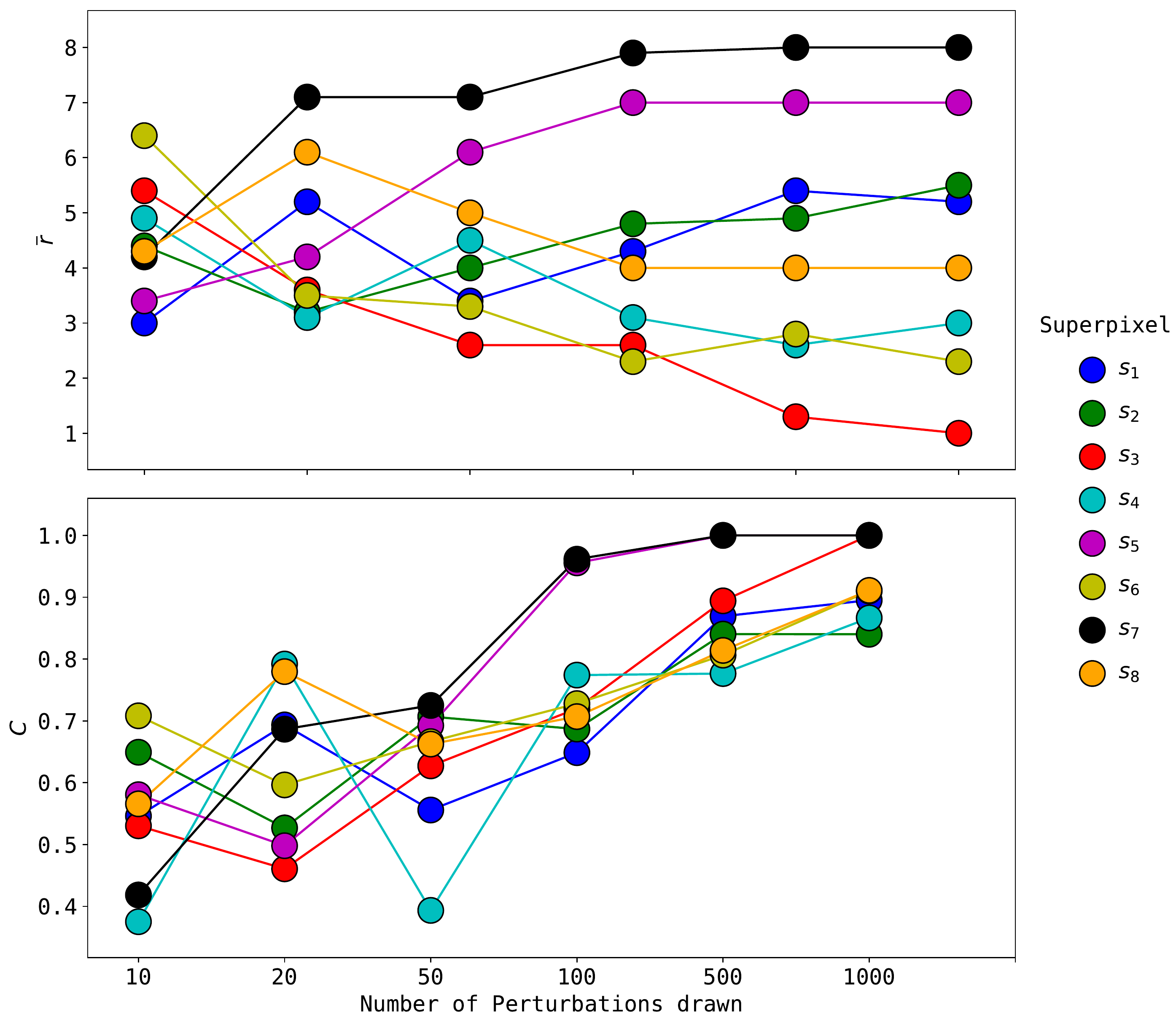}
  \caption{Influence of the number of perturbations}
  \label{c_over_r}
\end{subfigure}%
\newline
\begin{subfigure}{.5\textwidth}
  \centering
  \includegraphics[width=1\linewidth]{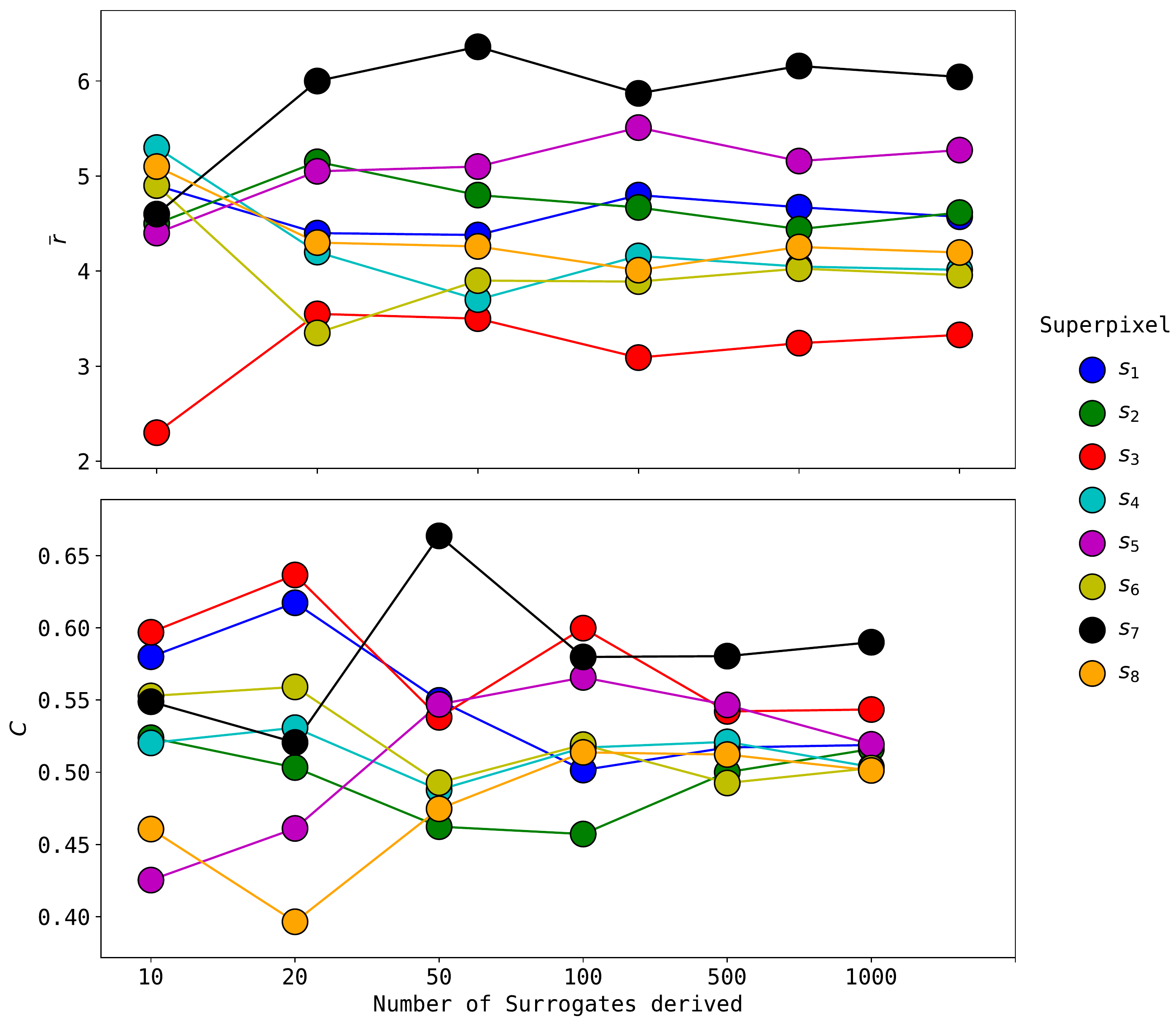}
  \caption{Influence of the number of surrogates}
  \label{c_over_r_boot}
\end{subfigure}
\caption{(a) 200 surrogates are derived on bootstrapped perturbations sets. For the superpixels corresponding to the example depicted in figure \ref{segmentationexample} (top row) mean rank $\overline{r}$ and ordinal consensus $C$ are plotted against the number of perturbations drawn for each surrogate. 
(b) 200 surrogates are derived on fixed perturbations sets of 200 samples. For the superpixels corresponding to the example depicted in figure \ref{segmentationexample} (top row) mean rank $\overline{r}$ and ordinal consensus $C$ are plotted against the number of surrogates derived from the perturbation dataset.} 
\label{C_over}
\end{figure}


\paragraph{Datasets}
For the image classification task we use the CIFAR-10 dataset \cite{cifar10} For this work, the data set is split to a training set and a validation set with 50000 and 10000 images respectively. For sentiment classification we use the movie review dataset IMDB\cite{imbd}. The task is to classify a movie review as positive or negative based on the given text. The dataset consists of 50000 labelled reviews. 

\paragraph{Models}
For our black-box image classifier we use an ensemble of 5 CNNs with ResNet  architecture as in \cite{resnet}. The ensemble is created by training all CNNs individually, using random weight initialisation \cite{glorotuniform} and data shuffling during training to induce diversity \cite{ensemble}. For the text analysis we use an ensemble of fully connected neural networks in combination with GloVe embeddings \cite{glove}, using random weight initialising and data shuffling during training to induce diversity among the ensemble members. The default configuration of LIME is used with linear regression as surrogates \cite{lime}.

 
\paragraph{Results}\label{results}

In Fig.~\ref{c_over_r} we see that by increasing the number of perturbations for each bootstrap sample, the mean ranking of the superpixels converges towards values on the full ranking interval 1 to 8, whereas for small numbers of perturbations, the mean ranks are squashed into a rather small interval (top plot). The bottom plot shows that the level of agreement $C$ of the ranking of superpixels increases as the number of image perturbations is also increased. This is expected as the surrogates are trained on datasets that are more similar between them. Therefore, the explanation derived by aggregating multiple surrogates on more perturbations can be considered more certain with regards to the individual superpixels ranking. Examining both plots depicted in fig. \ref{c_over_r}, the highest agreement for the highest and lowest ranked superpixels ($s_7$ and $s_3$) among the surrogates is maximised. In Figure \ref{c_over_r_boot}, the same experiment is run for different numbers of bootstrap samples for a fixed number of perturbations. We see that increasing the number of surrogates does not increase the agreement of the raters assigning ranks to the superpixels (bottom plot). Contrary to Figure \ref{c_over_r}, the ranks do not converge to their absolute ranking. The agreement measured by the consensus estimated $C$, however, changes showing an increasing or decreasing trend.
\begin{figure}[!t]
\centering
\begin{subfigure}{.46\textwidth}
  \centering
  \includegraphics[width=1\linewidth]{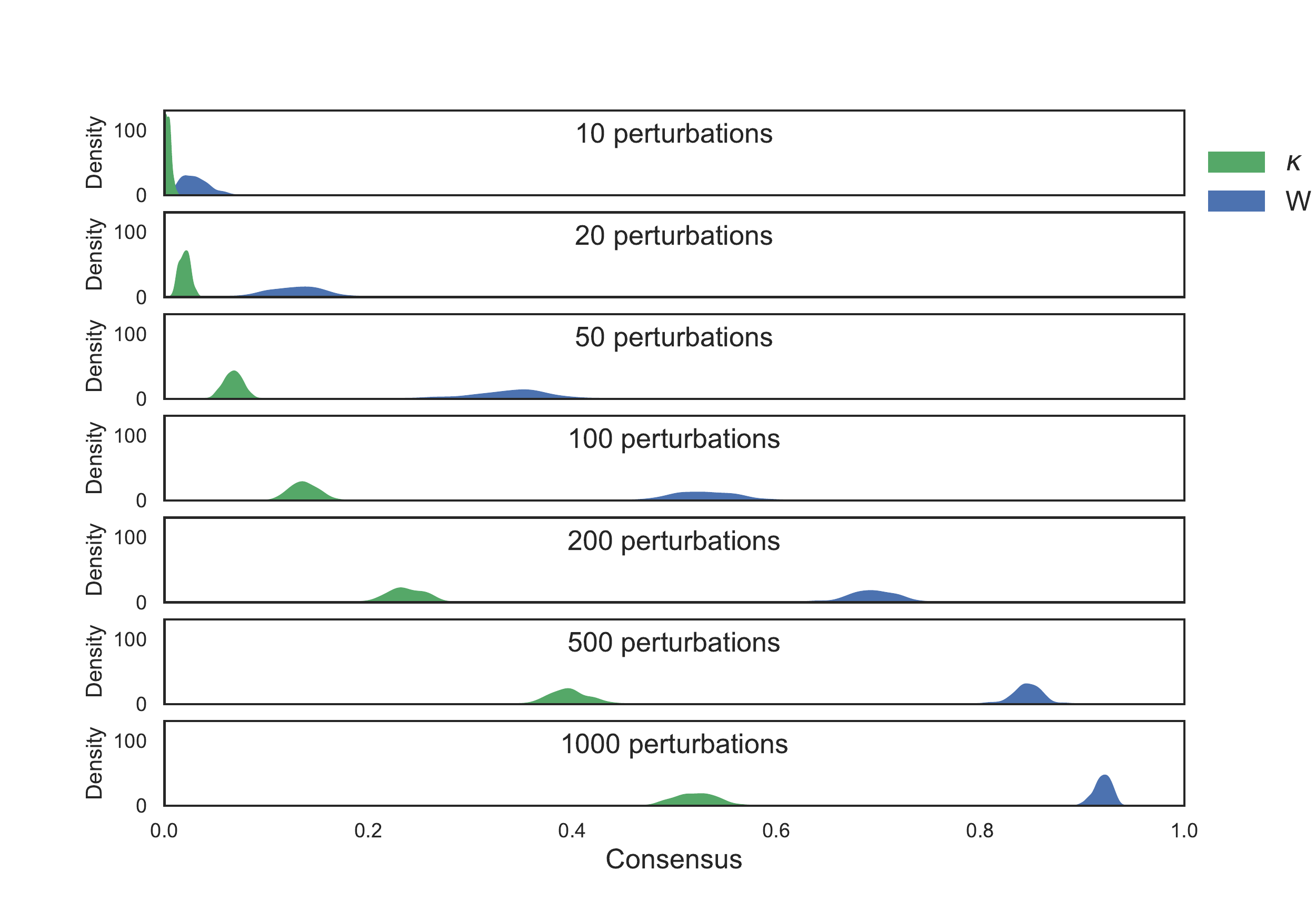}
  \caption{Influence of the number of perturbations}
  \label{W_over_pert}
\end{subfigure}%
\newline
\begin{subfigure}{.46\textwidth}
  \centering
  \includegraphics[width=1\linewidth]{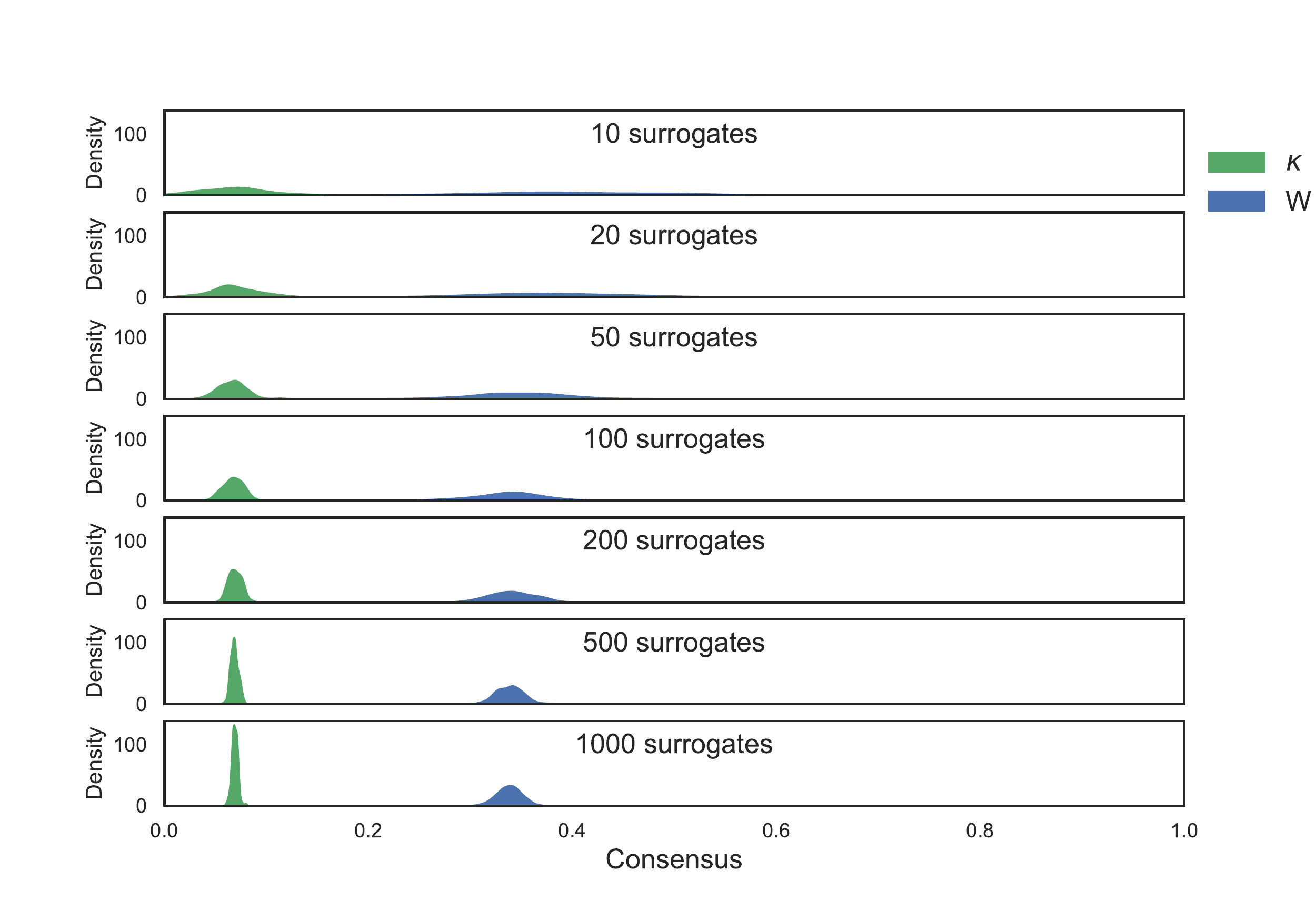}
  \caption{Influence of the number of surrogates}
  \label{W_over_surrogate}
\end{subfigure}
\caption{(a) 100 surrogates using bootstrapping on the perturbation dataset $\mathcal{P}$. We report the uncertainty estimates using the consensus $\kappa$ and $W$. The procedure is repeated 100 times with varying numbers of perturbations. 
(b) 100 sets $\mathcal{P}$ are drawn to fit the surrogates. We report the consensus estimates $\kappa$ and $W$. The procedure is repeated 100 times with varying numbers of surrogates.} 
\label{W_over}
\end{figure}

\begin{figure}[!t]
    \centering
    \includegraphics[width=0.85\columnwidth]{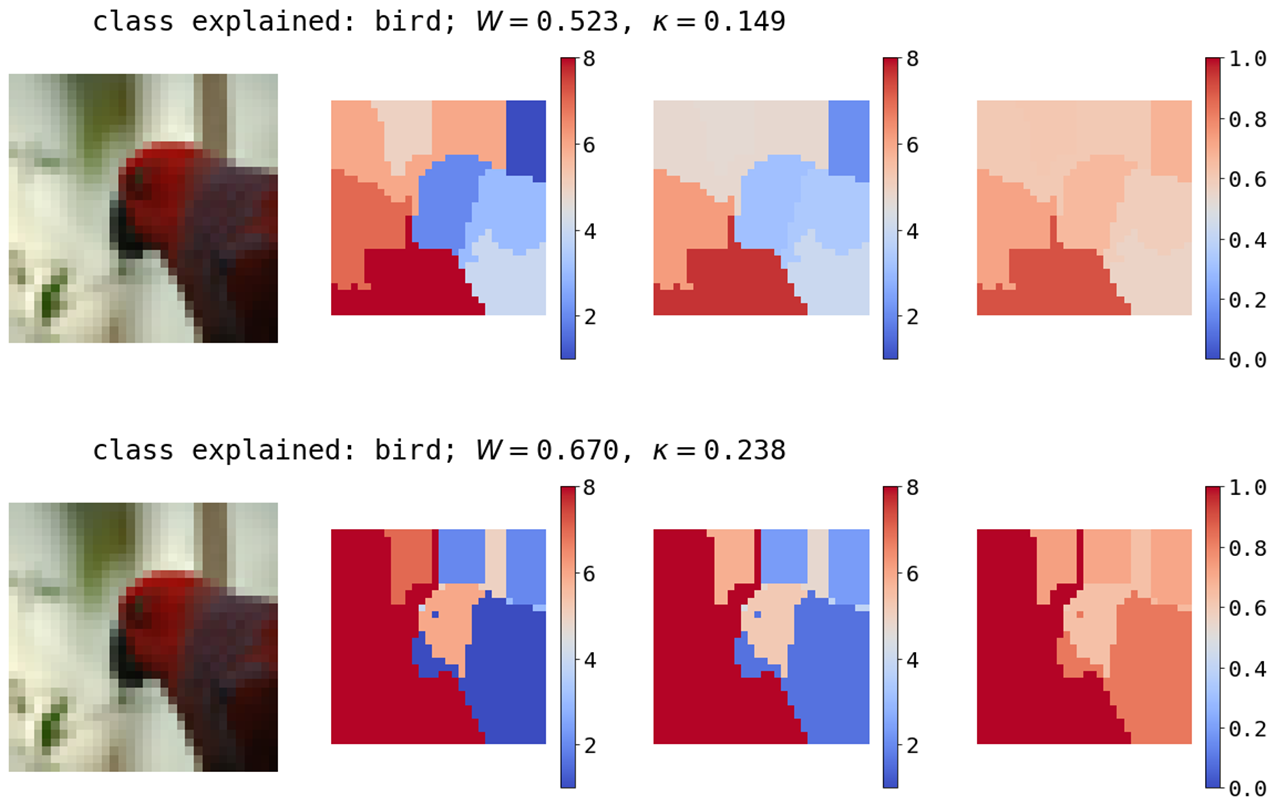}
     \caption{Left: original image $\mathbf{x}$, centre-left: absolute ranking of superpixels, centre-right: mean ranks, right: ordinal consensus.
    100 perturbation sets $\mathcal{P}_k$ drawn, 100 data points each to derive surrogates for the predicted class \textit{bird}.}
    \label{segmentationexample}
\end{figure}

\begin{figure}
    \centering
    \includegraphics[width=0.94\columnwidth]{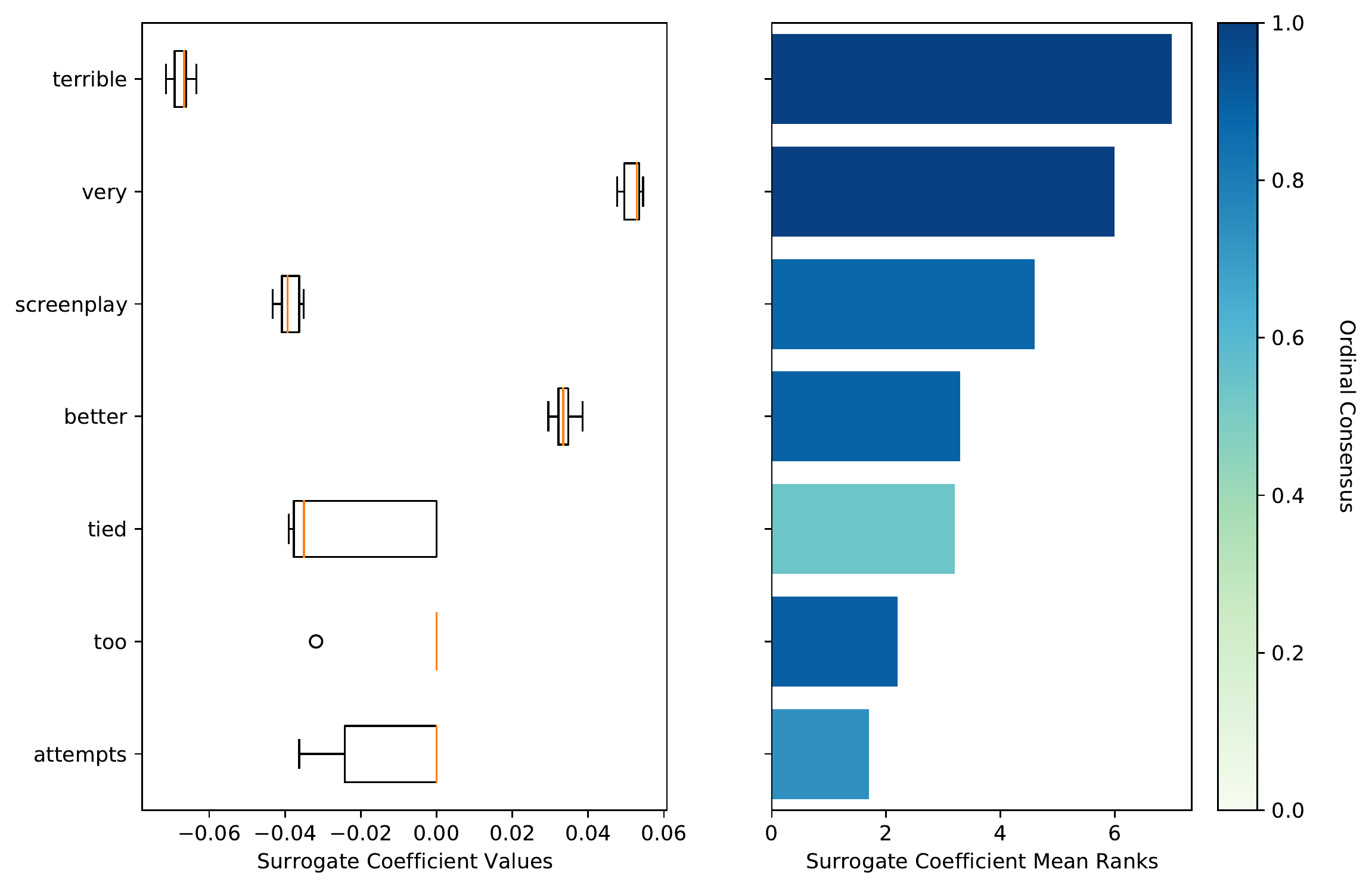}
    \caption{100 local explanations for a fixed data point using ensemble of 5 fully-connected NNs, 2 hidden layers, IMBD dataset. The mean rank $\overline{r}$ indicates the feature importance. 
    }
    \label{imbd}
\end{figure}

In Figs.~\ref{W_over_pert} and ~\ref{W_over_surrogate} we examine effects of the number of perturbations drawn and the number of surrogates derived on the uncertainty estimates $\kappa$ and $W$. As shown in Fig.~\ref{W_over_pert}, increasing the number of perturbations drawn shifts the distributions of consensus estimates towards higher values which matches the findings from Fig.~\ref{c_over_r}. Fig.~\ref{W_over_surrogate}, however, indicates that the distribution of derived uncertainty estimates become more narrow, resulting in more reliable estimates. 
Figures~\ref{segmentationexample} and ~\ref{imbd} show examples where our method provides additional information to the practitioner about the explanation. In fig.~\ref{segmentationexample} our method allows the user to compare different image segmentations for training surrogates. Absolute ranking of the mean ranks of the superpixels is shown in the second column. The mean ranks of superpixels are depicted in the third column. The image in the rightmost column shows the level of agreement amongst the surrogates regarding the ranking of the individual superpixels measured using the ordinal consensus $C$. The original image is segmented into 8 superpixels using different segmentation algorithms. By inspecting the uncertainty estimates $W$ and $\kappa$, the practitioner can conclude that the bottom segmentation results in an overall more certain explanation, as $W$ and $\kappa$ are higher compared to the segmentation in the top row. The example depicted in Fig.~\ref{imbd} shows our method on a text dataset (IMDB). Here, we highlight how the higher variance of surrogate coefficients is also shown by the ordinal consensus $C$.


\section{Conclusion}\label{discussion}
In this paper we make the case for the importance of reporting an uncertainty estimated of an explanation together with the explanation, when explaining a prediction. This  provides the user with the option of rejecting an explanation for being too uncertain. To this end, we proposed a procedure where we first bootstrap LIME, and then aggregate the outputs using ordinal statistics, obtaining both an explanation and a measure of its uncertainty.

\appendix

\newpage
\bibliographystyle{abbrv}
\bibliography{references}

\end{document}